\documentclass[letterpaper]{article} % DO NOT CHANGE THIS
\usepackage{aaai24}  % DO NOT CHANGE THIS
\usepackage{times}  % DO NOT CHANGE THIS
\usepackage{helvet}  % DO NOT CHANGE THIS
\usepackage{courier}  % DO NOT CHANGE THIS
\usepackage[hyphens]{url}  % DO NOT CHANGE THIS
\usepackage{graphicx} % DO NOT CHANGE THIS
\usepackage{natbib}  % DO NOT CHANGE THIS AND DO NOT ADD ANY OPTIONS TO IT
\usepackage{caption} % DO NOT CHANGE THIS AND DO NOT ADD ANY OPTIONS TO IT
\usepackage{algorithm}
\usepackage{algorithmic}

\usepackage{newfloat}
\usepackage{listings}
\DeclareCaptionStyle{ruled}{labelfont=normalfont,labelsep=colon,strut=off} % DO NOT CHANGE THIS
\floatstyle{ruled}
\newfloat{listing}{tb}{lst}{}
\floatname{listing}{Listing}
\title{DVQI: A Multi-task, Hardware-integrated Artificial Intelligence System for Automated Visual Inspection in Electronics Manufacturing}
\author {
    Audrey G. Chung\textsuperscript{\rm 1},
    Francis Li\textsuperscript{\rm 1},
    Jeremy Ward\textsuperscript{\rm 1},
    Andrew Hryniowski\textsuperscript{\rm 1,2,3},
    Alexander Wong\textsuperscript{\rm 1,2,3}
}
\affiliations {
    \textsuperscript{\rm 1}DarwinAI Corp., Waterloo, ON\\
    \textsuperscript{\rm 2}Vision and Image Processing Research Group, University of Waterloo\\
    \textsuperscript{\rm 3}Waterloo Artificial Intelligence Institute, Waterloo, ON\\
    $\{$audrey, francis, jeremy.ward, andrew, alex$\}$@darwinai.ca
}

\begin{document}

\maketitle

\begin{abstract}
As electronics manufacturers continue to face pressure to increase production efficiency amid difficulties with supply chains and labour shortages, many printed circuit board assembly (PCBA) manufacturers have begun to invest in automation and technological innovations to remain competitive. One such method is to leverage artificial intelligence (AI) to greatly augment existing manufacturing processes. In this paper, we present the DarwinAI Visual Quality Inspection (DVQI) system, a hardware-integration artificial intelligence system for the automated inspection of printed circuit board assembly defects in an electronics manufacturing environment. The DVQI system enables multi-task inspection via minimal programming and setup for manufacturing engineers while improving cycle time relative to manual inspection. We also present a case study of the deployed DVQI system's performance and impact for a top electronics manufacturer. 
\end{abstract}

\section{Introduction}
As electronics manufacturers aim to increase production efficiency amid difficulties with supply chains and labour shortages, many printed circuit board assembly (PCBA) manufacturers have begun to invest in automation and technological innovations to remain competitive. However, traditional automated inspection methods in the PCBA industry are costly to setup, and are difficult to extend to the later steps of the manufacturing process.

Printed circuit board assembly (PCBA) manufacturing is the process of placing and soldering electronic components on a printed circuit board (PCB). For a given board, this process is composed of many steps performed sequentially in a manufacturing line. While exact steps may vary, at a high level a typical line will consist of:
\begin{enumerate}
  \item Solder application: solder is applied on the bare PCB
  \item Pick and place: surface mount devices (SMD) are placed on the board
  \item Reflow oven: solder is melted and SMD components are soldered onto the board
  \item Through-hole assembly: through-hole components are placed onto the board 
  \item Wave solder: through-hole components are soldered onto the board 
  \item Final assembly: connectors, wires, additional mechanical components are assembled onto the board
  \item Box-build: if applicable, the board is assembled into its chassis 
\end{enumerate}
At any given step above defects may occur and make its way in to the final PCBA. Typically, a high-volume PCBA manufacturer would use an automated optical inspection (AOI) system right after the reflow oven to catch potential defects when placing and soldering the SMT components. While these AOIs can be effective at detecting defects, they mostly rely on traditional machine vision methods and hence are timely to setup and prone to false positives. This makes AOIs difficult to use in low-volume high-mix use cases, where different types of boards are frequently being made and the large overhead in programming the AOI per board becomes too costly. At the same time, AOIs are highly specialized in inspecting SMT components. The PCBA can become quite complex in the later stages after wave solder which the AOI is unable to inspect due to a large variation in electronic component height, occlusions, and positioning. 

\begin{figure}[t]
    \centering
    \includegraphics[width=\linewidth]{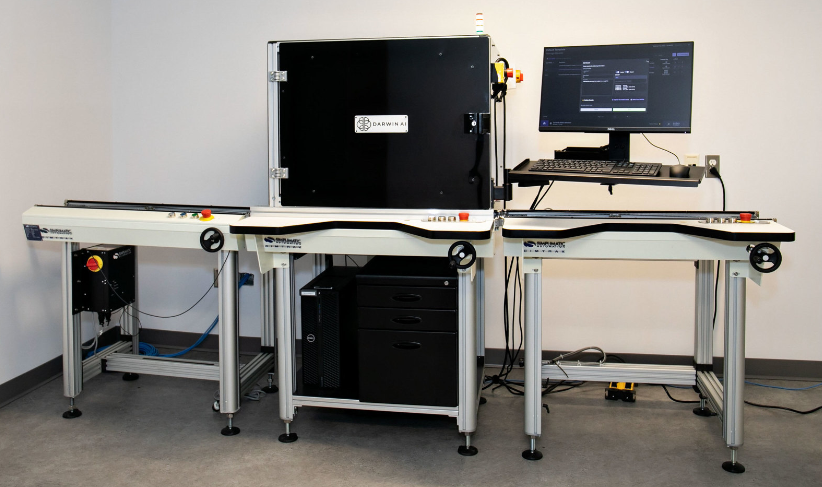}
    \caption{The proposed DVQI system in a conveyored configuration for inline inspection between two transport conveyors.}
    \label{fig_system}
\end{figure}

\begin{figure*}[t]
    \centering
    \includegraphics[width=\textwidth]{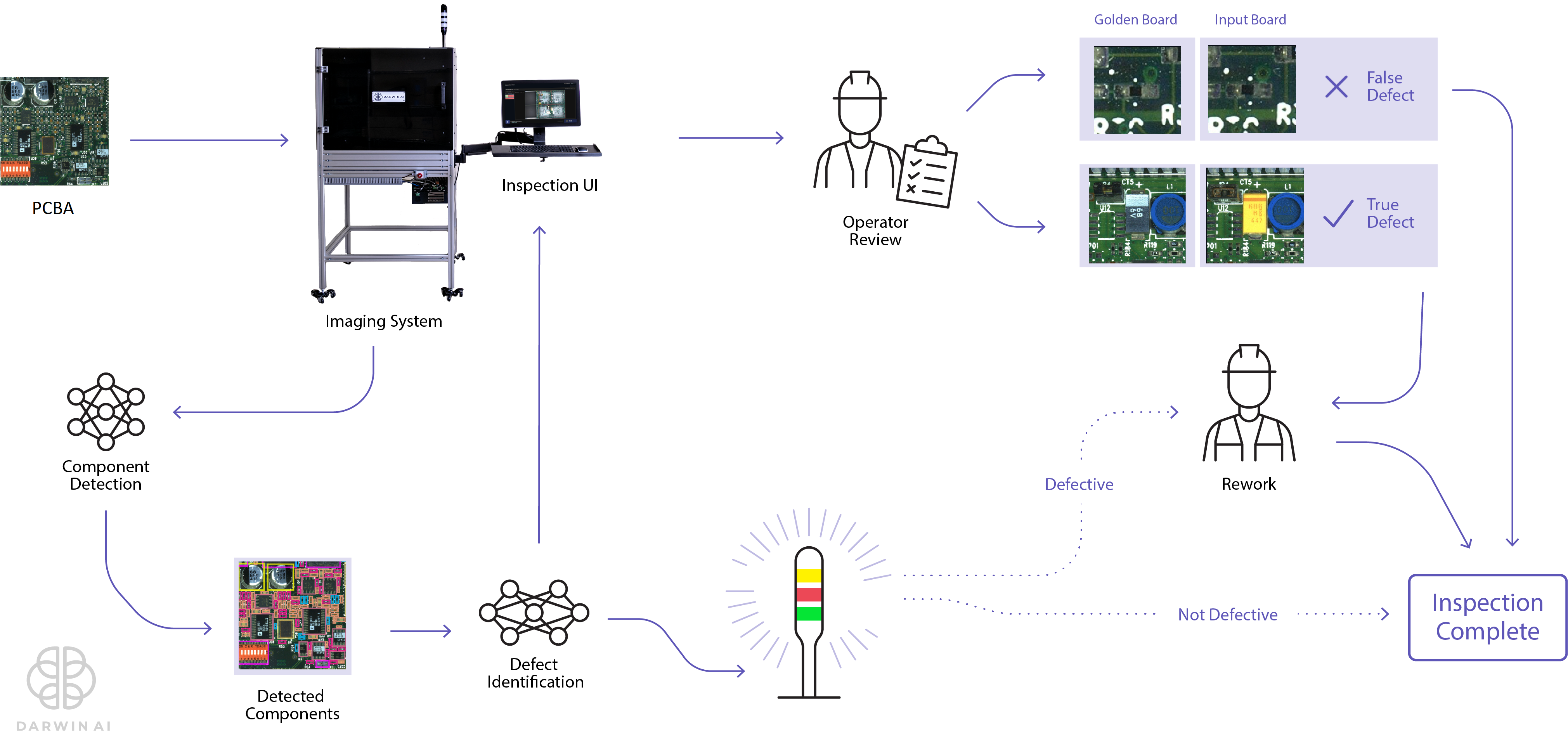}
    \caption{An overview of the inspection workflow using the proposed DarwinAI Visual Quality Inspection (DVQI) system.}
    \label{fig_algDesign}
\end{figure*}

As such, inspection for such high-variance, mixed-assembly PCBA scenarios are often conducted by human operators manually. However, not only is this time-consuming it is also prone to varying levels of inspection performance as the level of performance achieved by human inspectors can differ significantly~\cite{inspection1,inspection2}. A study by Sandia National Laboratories found that human inspectors miss 20\% to 30\% of defects across multiple types of inspection tasks~\cite{inspection3}.

For these reasons there is a need for a flexible system that can inspect mixed-assembly PCBAs that is both highly accurate while easy to setup. Recent advances in AI allows for such a system, where deep learning can be applied to help solve various defect detection challenges with PCBs and PCBAs~\cite{ling2023pcb,autoinsp1,autoinsp2,autoinsp3,autoinsp4,autoinsp5}. In this work we introduce DarwinAI Visual Quality Inspection (DVQI) system, a hardware-integrated artificial intelligence system which fulfills the needs of mixed-assembly PCBA inspection through automated multi-task visual inspection powered by deep learning. Figure~\ref{fig_system} shows the proposed DVQI system in a conveyored configuration between two transport conveyors.

\section{DVQI System}
The proposed DarwinAI Visual Quality Inspection (DVQI) system is a hardware-integrated artificial intelligence system designed to automatically perform multiple visual inspection tasks across a variety of printed circuit board assembly defects in an electronics manufacturing environment. Figure~\ref{fig_algDesign} presents an overview of the inspection workflow using the proposed DVQI system, where a PCBA is inspected for defects via component detection and defect identification, and presented back to an operator for review. 

More specifically, the DVQI system comprises an ultra-wide-field-of-view, extended depth-of-field imaging system that captures high-resolution images of assembled printed circuit boards, and leverages a suite of high-performance deep neural networks to conduct various automated inspection tasks. 

The DVQI system is configurable for both inline inspection (i.e., integrated into a PCBA manufacturing line) and stand-alone inspection (i.e., operated by an inspector asynchronous from the manufacturing lines). 

\begin{figure*}[t]
    \centering
    \begin{tabular}{p{0.47\linewidth} p{0.47\linewidth}}
	\includegraphics[width=\linewidth]{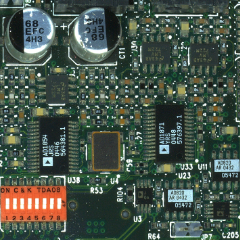}	&
	\includegraphics[width=\linewidth]{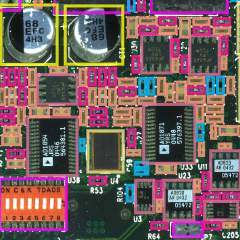} \\
    \end{tabular}
    \caption{The DVQI system takes image(s) of a PCBA to be inspected (left) and automatically detects and creates bounding boxes around every electronic component (right) as well as identify their component type via a highly efficient component detection deep neural network.}
    \label{fig_compDetects}
\end{figure*}

The DVQI system can be setup to inspect any assembled PCB configuration with minimal effort by simply inserting a ground-truth assembled PCB (referred to as a golden board) and initiating the automatic golden board learning process using a tailored component detection neural network (see the first subsection for details). Once set up, inspection can be performed on subsequent PCBAs by capturing high-resolution images of each board. These images, along with the reference images captured of the golden board, are then fed into a suite of customized deep neural networks trained on multiple inspection tasks to automatically compare the sample PCBAs against the golden board to detect and identify a plethora of possible defect types (see second subsection for details). These detected defects are then automatically logged and processed by the system for accountability and reporting purposes.

A human operator is then provided with alerts and visualizations of all detected defects (where the defects are, what type of defects they are, which board they belonged to, etc.) so that defective PCBAs can be scraped or reworked accordingly, and the manufacturing line can be adjusted to prevent further defects. The human operator is also given the option to provide active feedback to the DVQI system on whether it has made the right decisions. Doing so allows the system to continuously learn and improve over time as the system continues to be used (see the third subsection).

\subsection{Minimal Programming \& Setup}

\begin{figure}[t]
    \centering
	\includegraphics[width=\linewidth]{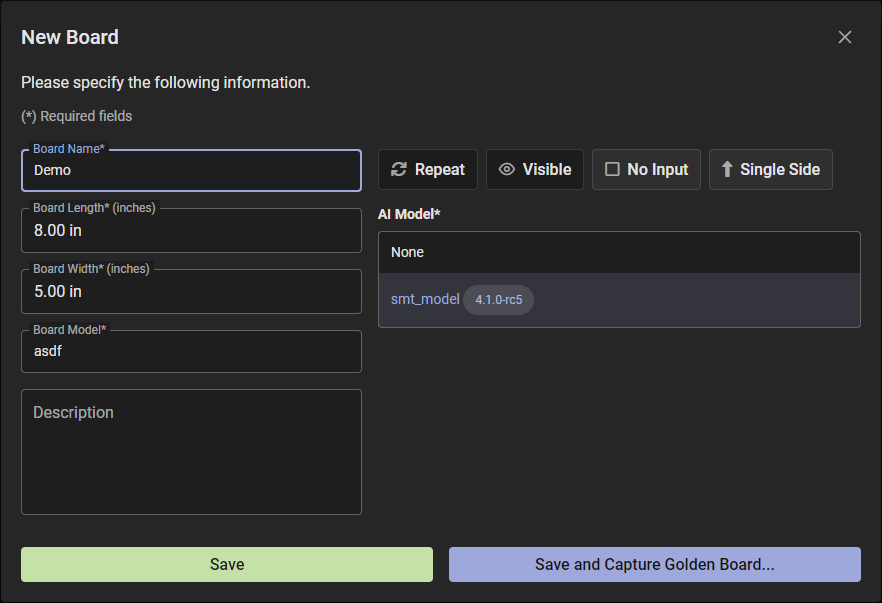}	
    \caption{The DVQI system minimizes the time required to program a new PCBA for inspection, and only requires the dimensions (length and width) of the PCBA during setup.}
    \label{fig_setup}
\end{figure}

In traditional visual inspection systems, each new PCBA design needs to be manually programmed into the system in order for the system to be able to perform visual inspection on that board.  More specifically, an operator needs to manually draw a bounding box around each electronic component on a golden board (a board with no defects), then program in a custom heuristic for determining whether that component is defective or not.  Given the presence of hundreds to thousands of components on each PCBA, this programming process can take as much as a day.  To minimize the need to program a new PCBA board type and go from a day to just a few minutes, the proposed DVQI system automatically learns a profile of a golden board and enables visual inspection immediately after via the following approach. As shown in Figure~\ref{fig_setup}, operators only need to input the dimensions (length and width) of the PCBA during the set up and programming of a new PCBA inspection. 

First, during the golden board learning process, an efficient deep neural network~\cite{li2023pcbdet} automatically detects and creates bounding boxes around every electronic component on the golden board.  More specifically, the neural network possesses a one-stage double-condensing attention condenser object detection architecture design achieved via a generative neural network design strategy, named generative synthesis~\cite{wong2019gensynth}, to automatically find the optimal balance between operational efficiency requirements for real-time, high-throughput inference and component detection accuracy.  We harness double-condensing attention condensers~\cite{wong2023faster} in the generative design strategy as they allow for highly condensed feature embeddings to achieve selective attention in a highly computational and efficient manner.  

Following automatic bounding box creation, a digital profile is created for each component automatically based on its component crop. This digital profile includes component details, such as component type (e.g., SMD resistor, integrated circuit, etc.), that are leveraged by the DVQI system to better constrain the search space of defects during multi-task inspection. The use of these digital profiles of components allows for the elimination of the need to manually create custom heuristics, while also significantly improving the robustness of the system as manual heuristics have significant challenges in handling a wide variety of variations in both defect and non-defect scenarios. 

\subsection{Multi-task Inspection}

\begin{figure}[t]
    \centering
    \begin{tabular}{p{0.47\linewidth} p{0.47\linewidth}}
	\includegraphics[width=\linewidth]{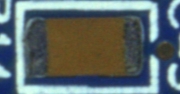}	&
	\includegraphics[width=\linewidth]{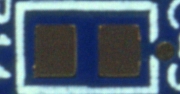} \\
    \includegraphics[width=\linewidth]{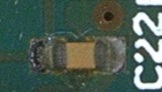}	&
	\includegraphics[width=\linewidth]{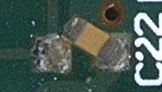} \\
    \includegraphics[width=\linewidth]{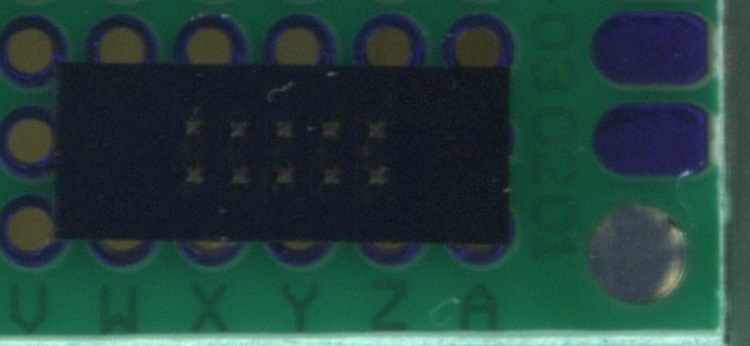}	&
	\includegraphics[width=\linewidth]{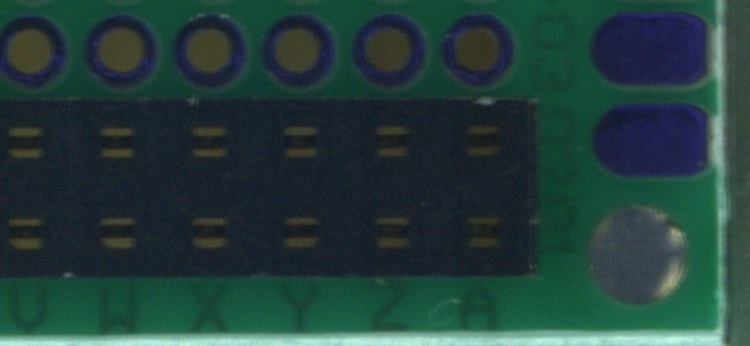} \\
    \includegraphics[width=\linewidth]{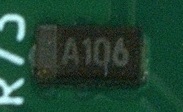}	&
	\includegraphics[width=\linewidth]{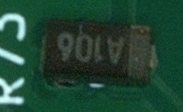} \\
    a) Reference component & b) Inspected component
    \end{tabular}
    \caption{Examples of the different defects that can be detected by the proposed DVQI system via a highly efficient multi-head deep neural network designed for multi-task inspection. From top to bottom: 1) presence / absence of a component, 2) skewed placement of a component, 3) wrong component used, and 4) reversed polarity placement of a component.}
    \label{fig_defects}
\end{figure}

During operational inspection the golden board for the appropriate PCBA is retrieved, and the digital profiles of all electronic components of the golden board are loaded to support the automatic visual inspection of the sample PCBA.  More specifically, an efficient multi-head, double-condensing attention condenser neural network, generated using the aforementioned generative network architecture design strategy~\cite{wong2019gensynth}, is leveraged to perform multiple inspection tasks on each electronic component to identify a variety of different types of defects.  Here, similar to the component detection network, we also harness double-condensing attention condensers~\cite{wong2023faster} in the generative design strategy to achieve selective attention to improve representational capabilities, which is quite important especially for multi-task applications, while achieving high computational and architectural efficiency.  Similar to \cite{wong2023fast} in terms of handling multiple tasks, the proposed network shares a common double-condensing attention condenser backbone for feature extraction, with each network head responsible for detecting a different defect: 1) presence / absence of a component, 2) skewed placement of a component, 3) wrong component used, and 4) reverse polarity placement of a component (see Figure~\ref{fig_defects} for examples of each defect). 

The key advantage of leveraging a deep neural network for automatically tackling such inspection tasks is that it not only eliminates the need for manual programming of individual components, but also allows for: 1) significant improvements in robustness during inspection as it can better handle high diversity and variations in defect and non-defect scenarios, 2) it allows for the identification of defect types for a given component (while heuristics only allow for identifying whether a defect occurs but not what type), and 3) it can be improved over time as new training data and new operator feedback on inspection decisions are collected. To train the deep neural networks within the DVQI system (component detection and multi-task inspection), we prepared a large-scale, carefully curated proprietary data comprising of around half a million fully annotated electronic components across a wide variety of component types and defect types. Currently, the DVQI system has an inspection accuracy of 97.8\%, with a false positive (or overkill) rate of 1.7\% and a false negative (or escape) rate of 3.7\%.

\begin{figure*}[t]
    \centering
    \includegraphics[width=\linewidth]{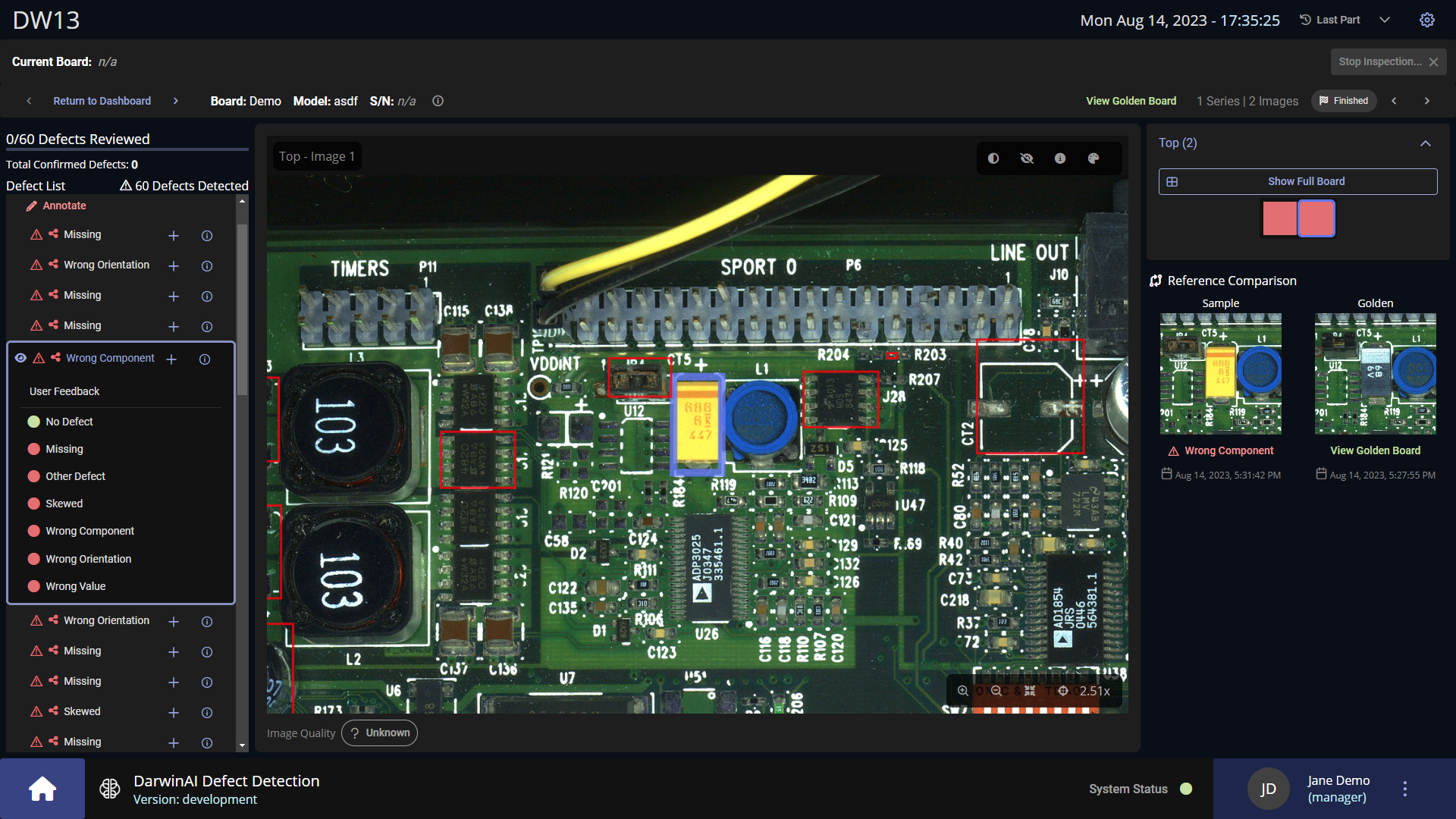}
    \caption{The DVQI system displays detected defects via a graphical user interface (GUI), allowing the operator to visualize and validate defects as necessary. The operator is also able to provide feedback on whether the deep neural network made the correct decision, and this information is used to improve the performance of the neural networks over time.}
    \label{fig_defect}
\end{figure*}

Once all the defect types and locations are automatically identified, they are displayed using a graphical user interface (GUI) for the operator to visualize and validate if needed. The DVQI system allows the operators to easily adjust the performance of the system during inspection by fine-tuning the sensitivity per inspected component, while also enabling the operators to provide feedback on whether the deep neural network made the correct decision and leverage this information to improve the performance of the neural networks over time.  Furthermore, the DVQI system also allows operators to add alternate components to standard components found on the golden board that may be used by the manufacturer during the PCBA manufacturing process. This is especially important in recent years, with manufacturers switching to different functionally-equivalent electronic components from different vendors due to supply chain issues.
This on-site continuous learning strategy leverages human-in-the-loop improvements that lead to the system performing increasingly better through use and allows the human operator to garner greater trust in the system over time.

\subsection{Ease-of-Use Considerations}
The DVQI system is first and foremost easy for operators to use. From setting up and programming the system to inspect a new PCBA to visualizing and validating defects in real-time to reviewing past inspections, operators can use the graphical user interface (GUI) to help streamline their inspection workflow. 

The DVQI system displays the inspection results and allows the operator to visualize and validate the detected defects as necessary. As shown in Figure~\ref{fig_defect}, a list of all detected defects is provided in the left panel of the GUI, detected defects are clearly indicated on the image(s) of the sample PCBA via red bounding boxes, and component crop comparisons against the reference component on the golden board (and any alternate components, if applicable) are available in the right panel for easy visual verification. The operator is also able to provide feedback on whether the deep neural network made the correct decision in the left panel, and this information is used to improve the performance of the neural networks over time.

\begin{figure*}[h!]
    \centering
	\includegraphics[width=\linewidth]{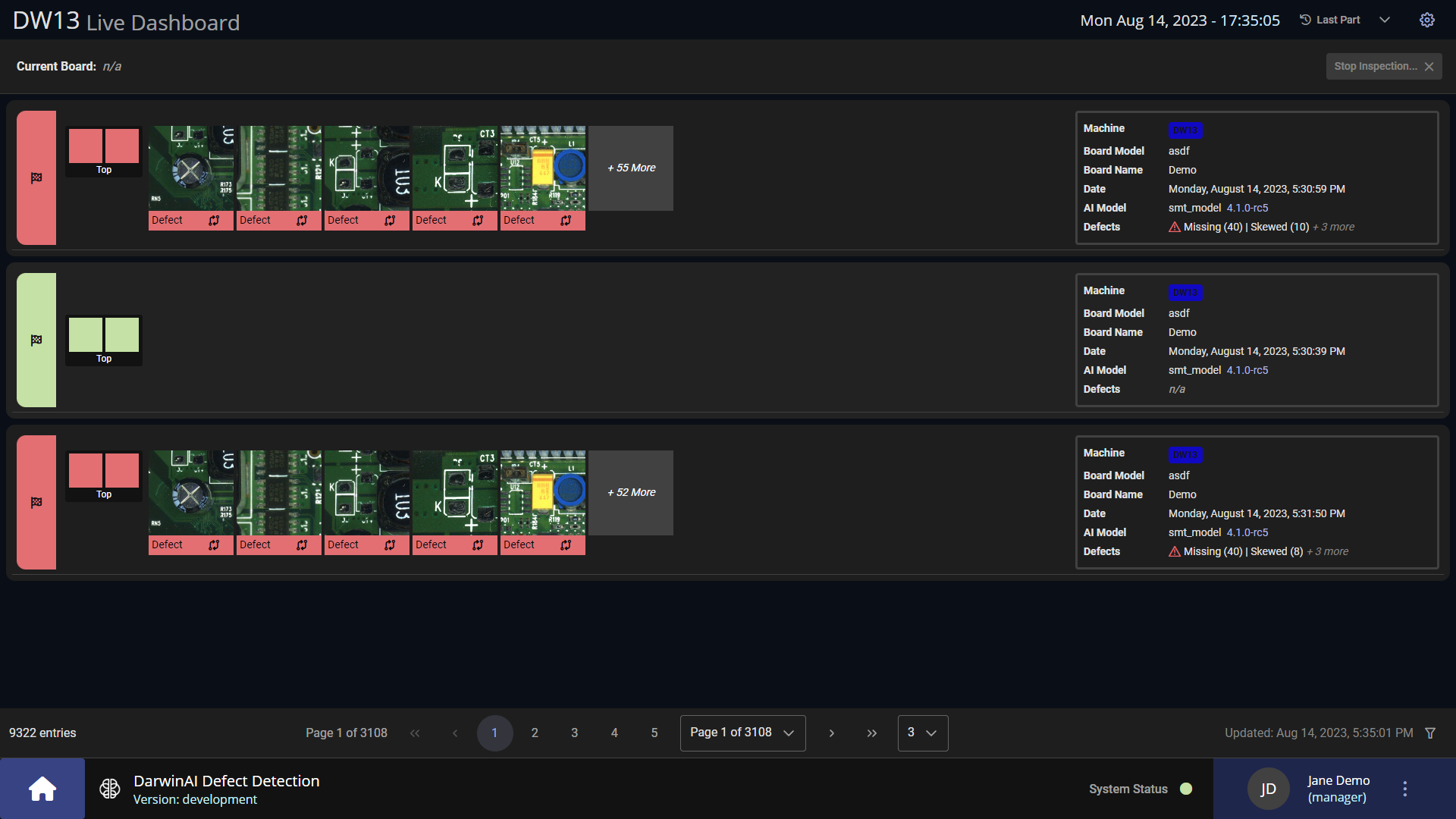}	\\
	\includegraphics[width=\linewidth]{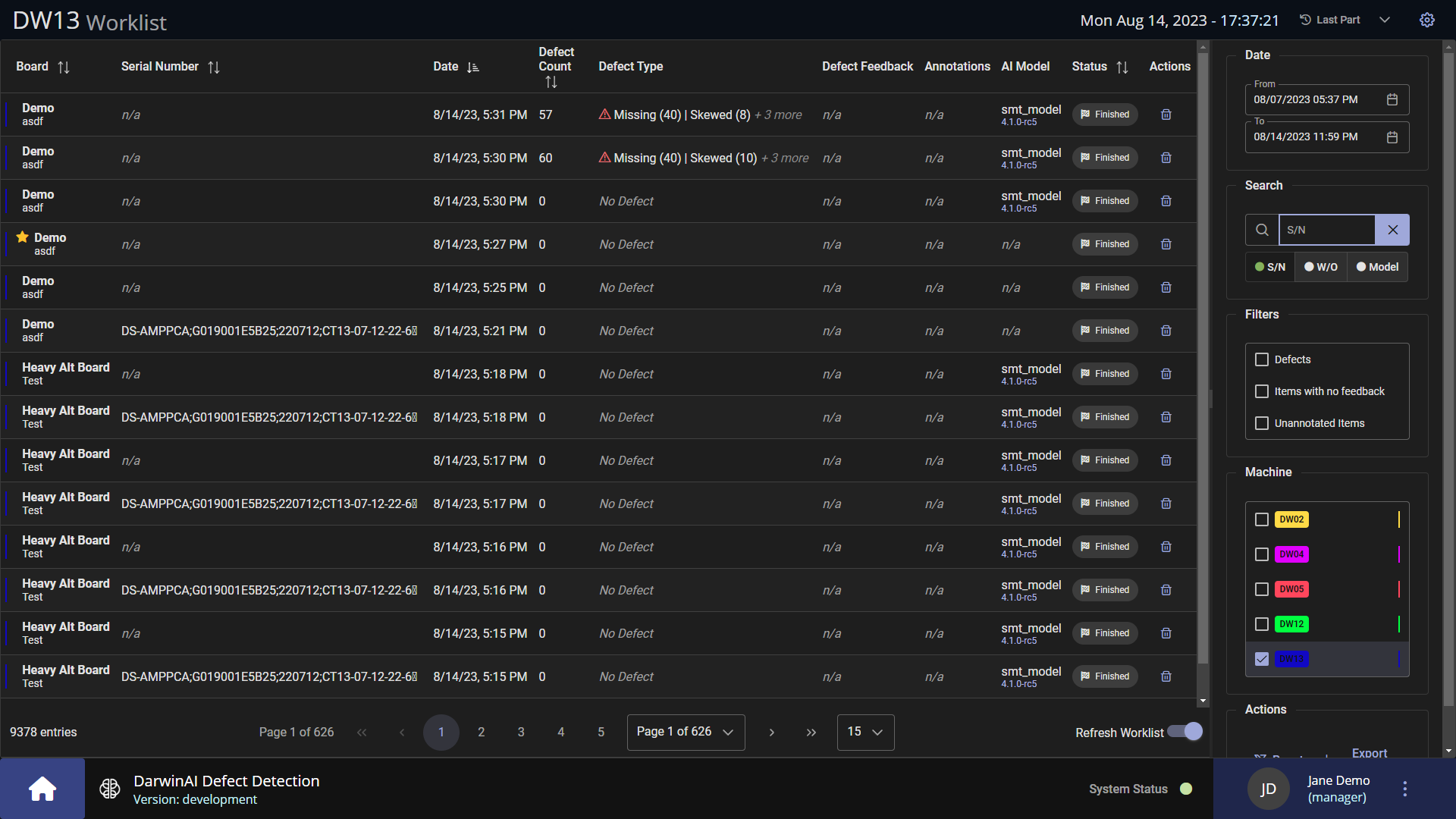} \\
    \caption{The DVQI system allows for operators to both easily monitor the manufacturing line in real-time via the live dashboard, and retroactively review past inspections via the worklist. The live dashboard (top) continually updates with the most recent inspection results, and provides an easy way for operators to see which boards contain defects as PCBAs are actively being inspected. The worklist (bottom) is used to easily review past inspection results, and has board serial number search and filtering capabilities.}
    \label{fig_monitor}
\end{figure*}

Lastly, the DVQI system allows for operators to both easily monitor the manufacturing line in real-time via the live dashboard and retroactively review past inspections via the worklist (see Figure~\ref{fig_monitor}). The live dashboard continually updates with the most recent inspection results, and provides an easy way for operators to see which boards contain defects via heuristic green (good, not defective) and red (bad, defective) colour-coding. The live dashboard also shows thumbnails of the detected defects for quick validation, and this view is typically used by operators as PCBAs are actively being inspected. 

The worklist displays a list view of all inspection results, and enables operators to quickly find and review past inspections. To help with this, the worklist allows the user to search by board serial number search, work order, or AI model, and has filtering capabilities (e.g., by date and time, manufacturing line, inspections that contain defects, etc.). This view is often most useful for users other than the operator who are not directly overseeing the manufacturing line and operating the inspection system, such as manufacturing engineers, rework technicians, and plant managers.

\subsection{Improved Cycle Time}
The DVQI system shows a clear improvement to inspection cycle time relative to the manual inspection of PCBAs. While manual inspection of PCBAs can take anywhere from a few minutes to over 10 minutes for more densely populated PCBAs. In comparison, the DVQI system's inspection cycle time is than one minute. When configured for stand-alone inspection, the inspection cycle time of the DVQI system is approximately 50 seconds, including the time spent physically loading and unloading the PCBA into the system. 

\begin{table}[b]
    \centering
    \begin{tabular}{|c|c|}
        \hline
        PCBA Size   &   Inspection Cycle Time \\ \hline
        8" x 8"     &   $\sim$ 20 s \\ 
        10" x 14"   &   $\sim$ 25 s \\
        20" x 20"   &   $\sim$ 40 s \\ \hline
    \end{tabular}
    \caption{Approximate inspection cycle times of the DVQI system when configured for inline inspection.}
    \label{tab_times}
\end{table}

During inline inspection, the DVQI system leverages standard SMEMA~\cite{ths2019ipc} communication signals via its conveyor base and the approximate inspection cycle times are shown in Table~\ref{tab_times}. In the case study presented in the following section, these inline inspection cycle times were sufficient to keep up with the manufacturing line whether the DVQI system is deployed. 

\section{Case Study: Deployment in Electronics Manufacturing}

We present a case study of the DarwinAI Visual Quality Inspection (DVQI) system's deployment at a top electronics manufacturer, where the system has been deployed for over a year. During the first 12 months of deployment, the DVQI system inspected approximately 55.5 million components across over 90k boards. The manufacturer has continued to use the DVQI system beyond the first year of deployment, and is actively using the system as part of their standard manufacturing process.

For the purpose of this paper, we specifically leverage the DVQI system performance over a one month evaluation period during the first year of deployment to provide a detailed analysis and summarize the impact of the system. During this evaluation period, the DVQI system was deployed at one of the manufacturer's surface-mount technology (SMT) lines, and inspected PCBAs coming out of a reflow oven for SMD component placement defects. An average of 388 boards were inspected daily and a total of 9,218 boards were inspected, with 473 defective boards caught over the course of the one month evaluation period.

\subsection{System Performance Across Deployed Environments}

During the evaluation period, a total of 5.7 million components were inspected for component placement defects, and the DVQI system achieved a false positive or overkill rate of 0.11\%. This is notably less than our internal false positive rate of 1.72\%, indicating that our continuous learning strategy of human-in-the-loop improvements can indeed allow for boosted inspection performance after the system has been deployed. 

We also assessed the availability and reliability (of the DVQI system in a deployed environment, as defined below: 
\[Availability\;(\%) = \frac{Uptime\;(min) - Downtime\;(min)}{Uptime\;(min)}\]
\[Reliability\;(hrs) = \frac{Total\;uptime}{\#\;of\;breakdowns}\] 
where uptime excludes any planned system maintenance and downtime refers to when the system is unavailable for inspection. Note that system reliability is measured as the mean time between failures (MTBF). During the evaluation period, the DVQI system achieved a system availability of 99.6\%, and a MTBF of 356 hrs.

\subsection{Impact for Manufacturers}
Lastly, we present the impact of the DVQI for the manufacturer during the one month evaluation period, and the anticipated return on investment (ROI) of using this system for a full year. The manufacturer saw notable benefits both in terms of first piece savings via labour costs of manual inspection, and cost avoidance via reducing waste and rework.

In terms of first piece savings, the manufacturer estimated an average of 15 first piece inspections per day, each taking 15 minutes and resulting in 225 minutes of manual inspection at a labour rate of \$33.78/hr. Over the course of a year, the manufacturer anticipated \$32,760 of savings from labour costs associated with manually inspecting the first piece of each PCBA batch. Similarly, the manufacturer estimated \$4,769 of cost avoidance by reducing waste and rework (i.e., cost saved by preventing multiple defects from being manufactured) per month, and anticipated \$56,228 in cost avoidance over the course of a year. Together, the manufacturer estimated \$89,988/year in added value by using the DVQI system on a single manufacturing line. 

\section{Discussion}
In this work, we introduced the DarwinAI Visual Quality Inspection (DVQI) system, a hardware-integration artificial intelligence system for the automated inspection of printed circuit board assembly defects in an electronics manufacturing environment. The DVQI system was designed to conduct automatic multi-task inspection of PCB boards for a multitude of different electronic component defects in an accurate yet high-throughput manner via a highly efficient double-condensing attention condenser network created via generative network architecture search.  The system was further designed to require minimal operator programming and setup through the use of a highly efficient deep neural network also created via generative network architecture search.  Finally, DVQI provides easy-to-use GUI mechanisms to facilitate operator review of detected defects and user feedback that enables continuous learning to be conducted to improve inspection performance by the deep neural networks over time, allowing the system to adapt to the manufacturing environment and operational needs.

A case study was conducted to investigate and evaluate the DVQI system in a deployment scenario at a top electronics manufacturer, which at the time of this paper has now inspected approximately 55.5 million electronic components across over 90k boards.  A number of key findings were observed during the case study.  First, the DVQI system was qualitatively evaluated to be intuitive to the user and easy to use to setup boards, taking on average 30 minutes or less to program the system to perform a new PCBA inspection. As well, the defect detection models were robust to small variations in the boards and generalized to components never seen before by the deep neural networks during training. Lastly, the deployed DVQI system has a lower false positive rate (0.11\%) than the default DVQI system performance (false positive rate of 1.7\%), indicating that the system's inspection performance improves over time via user feedback during deployment. 

Future system improvements include adding additional defect types to be detected by the system by extending the multi-task deep neural network, such as solder joint inspection and foreign object detection. 
In summary, it was observed that the DVQI system provides significant benefits within a deployed scenario, saving an estimated \$90k per manufacturing line per year in labour costs and prevented waste and rework, and enabling electronics manufacturers to increase production efficiency and remain competitive.

\appendix

\bibliography{DVQI}

\begin{thebibliography}{14}
\providecommand{\natexlab}[1]{#1}

\bibitem[{Bhatt et~al.(2021)Bhatt, Malhan, Rajendran, Shah, Thakar, Yoon, and
  Gupta}]{autoinsp4}
Bhatt, P.~M.; Malhan, R.~K.; Rajendran, P.; Shah, B.~C.; Thakar, S.; Yoon,
  Y.~J.; and Gupta, S.~K. 2021.
\newblock {Image-Based Surface Defect Detection Using Deep Learning: A Review}.
\newblock \emph{Journal of Computing and Information Science in Engineering},
  21(4).

\bibitem[{Kim et~al.(2021)Kim, Ko, Choi, and Kim}]{autoinsp1}
Kim, J.; Ko, J.; Choi, H.; and Kim, H. 2021.
\newblock {{P}rinted {C}ircuit {B}oard {D}efect {D}etection {U}sing {D}eep
  {L}earning via {A} {S}kip-{C}onnected {C}onvolutional {A}utoencoder}.
\newblock \emph{Sensors (Basel)}, 21(15).

\bibitem[{Klamklay and Bishu(1998)}]{inspection2}
Klamklay, J.; and Bishu, R.~R. 1998.
\newblock Visual Inspection of Circuit Boards: Effect of Gender, Age, Defect
  type, and Defect Proportion.
\newblock \emph{Proceedings of the Human Factors and Ergonomics Society Annual
  Meeting}, 42(16): 1161--1164.

\bibitem[{Li et~al.(2022)Li, Palayew, Li, Abbasi, Nair, and
  Wong}]{li2023pcbdet}
Li, B.; Palayew, S.; Li, F.; Abbasi, S.; Nair, S.; and Wong, A. 2022.
\newblock PCBDet: An Efficient Deep Neural Network Object Detection
  Architecture for Automatic PCB Component Detection on the Edge.

\bibitem[{Ling and Isa(2023)}]{ling2023pcb}
Ling, Q.; and Isa, N. A.~M. 2023.
\newblock Printed Circuit Board Defect Detection Methods Based on Image
  Processing, Machine Learning and Deep Learning: A Survey.

\bibitem[{See(2021)}]{inspection3}
See, J.~E. 2021.
\newblock Crowdsourcing Visual Inspection (No. SAND2021-11446C).
\newblock \emph{Sandia National Lab.(SNL-NM)}, Albuquerque, NM (United States).

\bibitem[{See et~al.(2017)See, Drury, Speed, Williams, and
  Khalandi}]{inspection1}
See, J.~E.; Drury, C.~G.; Speed, A.; Williams, A.; and Khalandi, N. 2017.
\newblock The Role of Visual Inspection in the 21st Century.
\newblock \emph{Proceedings of the Human Factors and Ergonomics Society Annual
  Meeting}, 61(1): 262--266.

\bibitem[{{The Hermes Standard Initiative} et~al.(2019)}]{ths2019ipc}
{The Hermes Standard Initiative}; et~al. 2019.
\newblock IPC-HERMES-9852: The global standard for machine-to-machine
  communication in SMT assembly (version 1.2).
\newblock Technical report, Technical report, IPC.

\bibitem[{Westphal and Seitz(2021)}]{autoinsp3}
Westphal, E.; and Seitz, H. 2021.
\newblock A machine learning method for defect detection and visualization in
  selective laser sintering based on convolutional neural networks.
\newblock \emph{Additive Manufacturing}, 41: 101965.

\bibitem[{Wong et~al.(2023{\natexlab{a}})Wong, Shafiee, Abbasi, Nair, and
  Famouri}]{wong2023faster}
Wong, A.; Shafiee, M.~J.; Abbasi, S.; Nair, S.; and Famouri, M.
  2023{\natexlab{a}}.
\newblock Faster Attention Is What You Need: A Fast Self-Attention Neural
  Network Backbone Architecture for the Edge via Double-Condensing Attention
  Condensers.

\bibitem[{Wong et~al.(2019)Wong, Shafiee, Chwyl, and Li}]{wong2019gensynth}
Wong, A.; Shafiee, M.~J.; Chwyl, B.; and Li, F. 2019.
\newblock GenSynth: a generative synthesis approach to learning generative
  machines for generate efficient neural networks.

\bibitem[{Wong et~al.(2023{\natexlab{b}})Wong, Wu, Abbasi, Nair, Chen, and
  Shafiee}]{wong2023fast}
Wong, A.; Wu, Y.; Abbasi, S.; Nair, S.; Chen, Y.; and Shafiee, M.~J.
  2023{\natexlab{b}}.
\newblock Fast GraspNeXt: A Fast Self-Attention Neural Network Architecture for
  Multi-task Learning in Computer Vision Tasks for Robotic Grasping on the
  Edge.

\bibitem[{Yang et~al.(2020)Yang, Li, Wang, Dong, Wang, and Tang}]{autoinsp5}
Yang, J.; Li, S.; Wang, Z.; Dong, H.; Wang, J.; and Tang, S. 2020.
\newblock Using Deep Learning to Detect Defects in Manufacturing: A
  Comprehensive Survey and Current Challenges.
\newblock \emph{Materials}, 13(24).

\bibitem[{Zhang et~al.(2022)Zhang, Zhang, Gamanayake, Yuen, Geng, Jayasekara,
  wei Woo, Low, Liu, and Guan}]{autoinsp2}
Zhang, Q.; Zhang, M.; Gamanayake, C.; Yuen, C.; Geng, Z.; Jayasekara, H.; wei
  Woo, C.; Low, J.; Liu, X.; and Guan, Y.~L. 2022.
\newblock {Deep learning based solder joint defect detection on industrial
  printed circuit board X-ray images}.
\newblock \emph{Complex \& Intelligent Systems}, 8: 1525--1537.

\end{thebibliography}

\end{document}